\documentclass[11pt,a4paper]{article}
\usepackage[hyperref]{emnlp2020}
\usepackage{times}
\usepackage{latexsym}
\usepackage{pgfplots}

\usepackage{microtype}
\usepackage{graphicx}
\usepackage{amsmath,amssymb}
\usepackage{booktabs}
\usepackage{multirow}
\usepackage[normalem]{ulem}
\usepackage{framed}
\usepackage{tabu}

\graphicspath{{figures/}}

\def\figref#1{Fig.~\ref{#1}}
\def\secref#1{Sec.~\ref{#1}}
\def\tabref#1{Table~\ref{#1}}

\aclfinalcopy 

\newcommand\dataset{\textsc{HoVer}}
\newcommand\datasize{26k}

\title{\dataset{}: A Dataset for Many-Hop Fact Extraction And Claim Verification}

\author{
 Yichen Jiang$^{\dagger}$\thanks{\:\:Equal contribution.} \ \ \ \ \ \ \ Shikha Bordia$^{\ddagger}$\footnotemark[1] \ \ \ \ \ \ \  Zheng Zhong$^{\ddagger}$ \ \ \ \ \ \ \  Charles Dognin$^{\ddagger}$ \\ \textbf{Maneesh Singh}$^{\ddagger}$ \ \ \ \ \ \ \  \textbf{Mohit Bansal}$^{\dagger}$ \vspace{5pt} \\
 $^{\dagger}$UNC Chapel Hill~~~~~$^{\ddagger}$Verisk Analytics, Inc. \\
 \small\texttt{\{shikha.bordia, zheng.zhong, charles.dognin, msingh\}@verisk.com}\\
 \small\texttt{ \{yichenj, mbansal\}@cs.unc.edu}
 }

\date{}

\begin{document}
\maketitle

\begin{abstract} 

We introduce \dataset{} (\textsc{Ho}ppy \textsc{Ver}ification), a dataset for many-hop evidence extraction and fact verification.
It challenges models to extract facts from several Wikipedia articles that are relevant to a claim and classify whether the claim is \textsc{Supported} or \textsc{Not-Supported} by the facts.
In \dataset{}, the claims require evidence to be extracted from as many as four English Wikipedia articles and embody reasoning graphs of diverse shapes. 
Moreover, most of the 3/4-hop claims are written in multiple sentences, which adds to the complexity of understanding long-range dependency relations such as coreference.
We show that the performance of an existing state-of-the-art semantic-matching model
degrades significantly on our dataset as the number of reasoning hops increases, hence demonstrating the necessity of many-hop reasoning to achieve strong results.
We hope that the introduction of this challenging dataset and the accompanying evaluation task will encourage research in many-hop fact retrieval and information verification.\footnote{We make the \textsc{HoVer} dataset publicly available at \\ \url{https://hover-nlp.github.io}}
\end{abstract}
\section{Introduction}
 
 \begin{table*}[!t]
\centering
\begin{small}
\begin{tabular}{p{0.2cm}|p{2.2cm}p{12.3cm}}
\toprule
\#H & Reasoning Graph & Examples\\
\midrule
  \multirow{4}{*}{2}& \multirow{4}{*}{\includegraphics[scale=0.15]{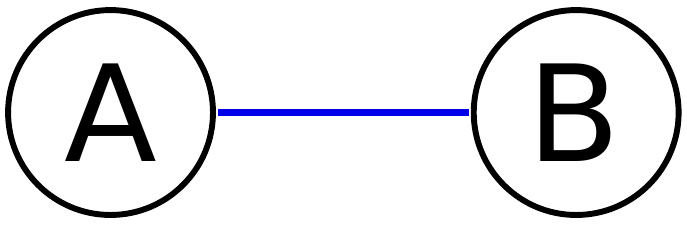}} &  \textbf{Claim}: Patrick Carpentier currently drives a Ford Fusion, introduced for model year 2006, in the NASCAR Sprint Cup Series. \newline
  \textbf{Doc A:} Ford Fusion is manufactured and marketed by Ford. Introduced for the 2006 model year, ...
  \textbf{Doc B:} Patrick Carpentier competed in the NASCAR Sprint Cup Series, driving the Ford Fusion.\\
\midrule
  \multirow{4}{*}{3}& \multirow{4}{*}{\includegraphics[scale=0.15]{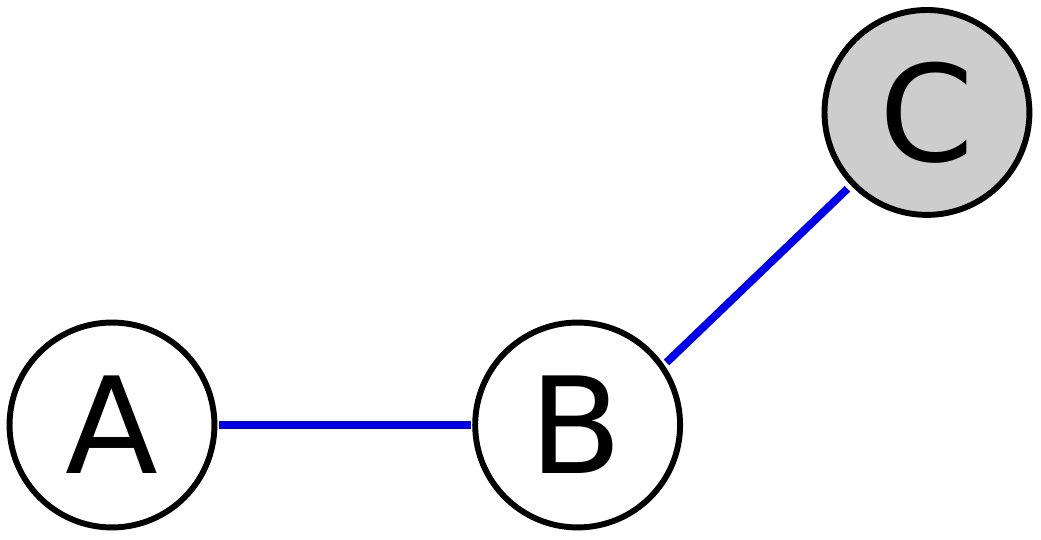}} & \textbf{Claim}: The Ford Fusion was introduced for model year 2006. \textit{The Rookie of The Year in the 1997 CART season} drives it in the NASCAR Sprint Cup Series. \newline 
  \textbf{Doc C:} The 1997 CART PPG World Series season, the nineteenth in the CART era of U.S. open-wheel racing, consisted of 17 races, ... Rookie of the Year was \textcolor{blue}{\underline{Patrick Carpentier}}.\\
\midrule
  \multirow{13}{*}{4}& \multirow{4}{*}{\includegraphics[scale=0.15]{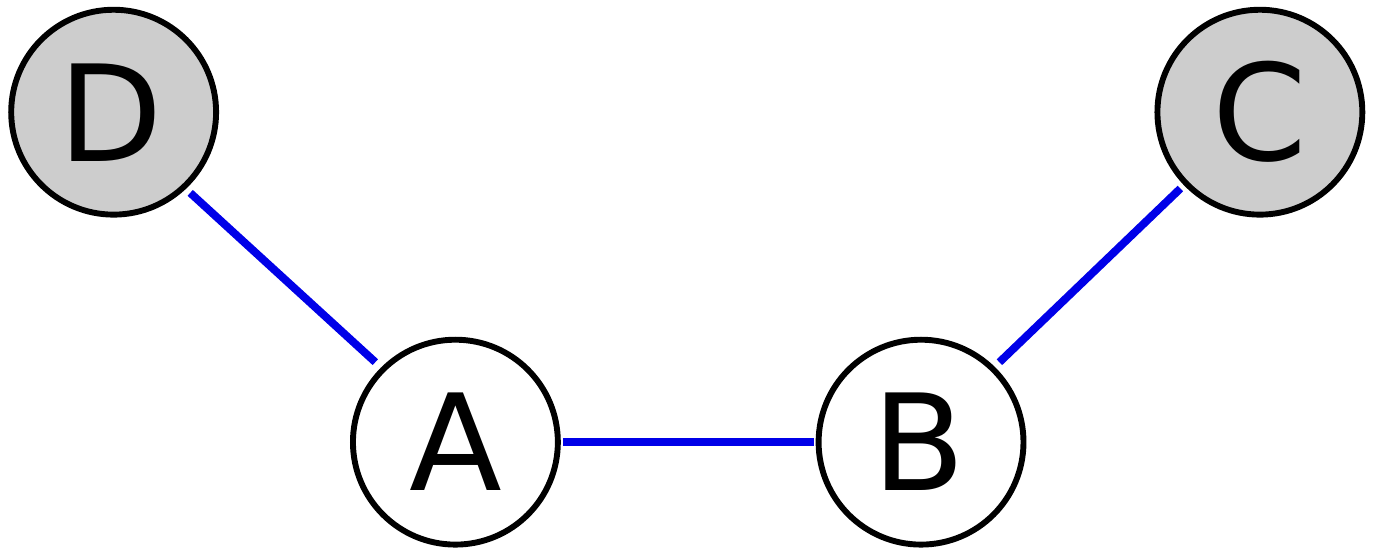}} & \textbf{Claim}: 
  \textit{The model of car Trevor Bayne drives} was introduced for model year 2006. The Rookie of The Year in the 1997 CART season drives it in the NASCAR Sprint Cup.\newline 
  \textbf{Doc D:} Trevor Bayne is an American professional stock car racing driver. He last competed in the NASCAR Cup Series, driving the No. 6 \textcolor{blue}{\underline{Ford Fusion}}...\\\cmidrule(lr){2-3}
  
  &\multirow{4}{*}{\includegraphics[scale=0.15]{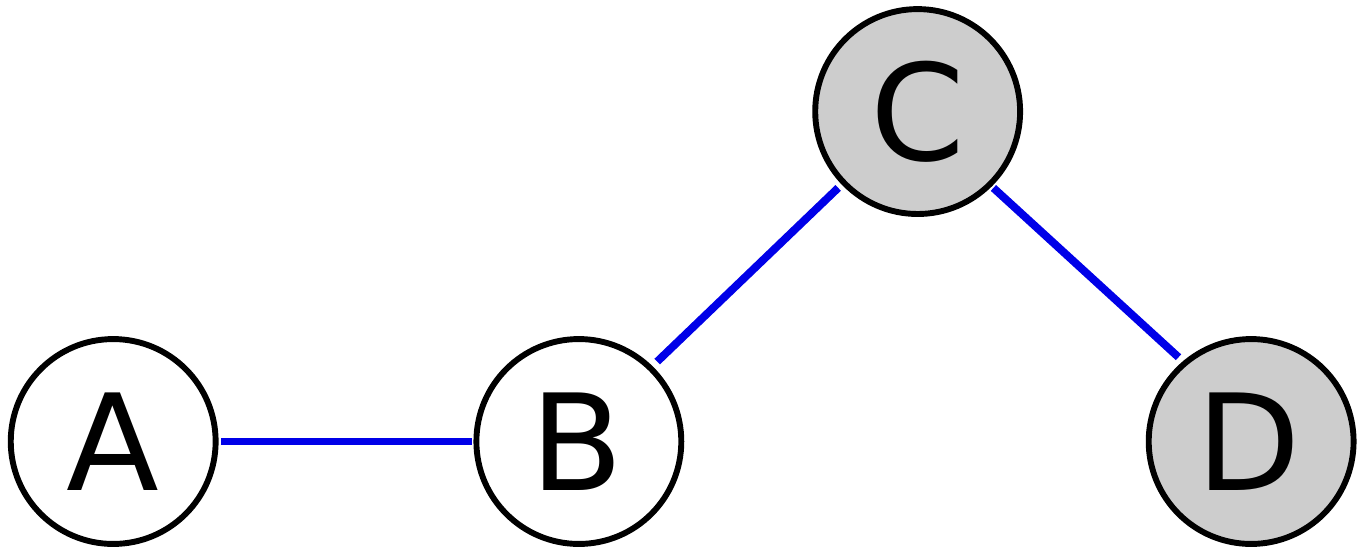}} & \textbf{Claim}: The Ford Fusion was introduced for model year 2006. It was driven in the NASCAR Sprint Cup Series by The Rookie of The Year of \textit{a Cart season, in which the 1997 Marlboro 500 was the 17th and last round}.\newline 
  \textbf{Doc D:} The 1997 Marlboro 500 was the 17th and last round of the \textcolor{blue}{\underline{1997 CART season}}... \\\cmidrule(lr){2-3}
  
  &\multirow{4}{*}{\includegraphics[scale=0.15]{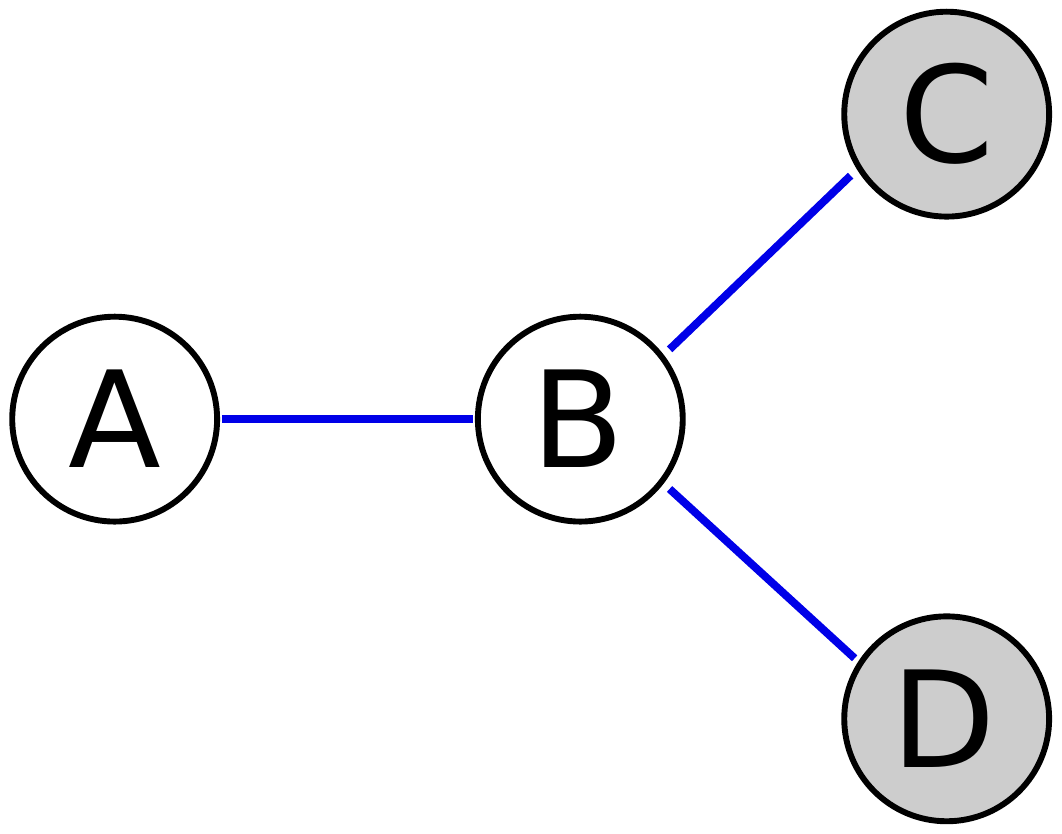}} & \textbf{Claim}: The Ford Fusion was introduced for model year 2006. The Rookie of The Year in the 1997 CART season drives it in the series held by \textit{the group that held an event at the Saugus Speedway}. \newline 
  \textbf{Doc D:} Saugus Speedway is a 1/3 mile racetrack in Saugus, California on a 35 acre site. The track hosted one \textcolor{blue}{\underline{NASCAR}} Craftsman Truck Series event in 1995... \\
\bottomrule
\end{tabular}
\end{small}
\vspace{-5pt}
\caption{Types (graph shape) of many-hop reasoning required to extract the evidence and to verify the claim in the dataset.
All claims presented are created and extended based on a single Q-A pair in \textsc{HotpotQA}.
The highlighted (blue+underlined) words from the original 2/3-hop claims are replaced with the italicized phrase based on the information from the newly-introduced \textbf{Docs} to form the 3/4-hop claims.
 \vspace{-5pt}
}
\label{table:reasoning-graphs}
\end{table*}

The proliferation of social media platforms and digital content has been accompanied by a rise in 
deliberate disinformation and hoaxes, leading to polarized opinions among masses. 
With the increasing number of inexact statements, there is a large interest in a fact-checking system that can verify claims based on automatically retrieved facts and evidence.
FEVER~\citep{thorne2018fever} is an open-domain fact extraction and verification dataset closely related to this real-world application.
However, more than 87\% of the claims in FEVER require information from a single Wikipedia article, while real-world ``claims" might refer to information from multiple sources.
QA datasets like \textsc{HotpotQA} \citep{yang2018hotpotqa} and QAngaroo \citep{welbl2018constructing} represent the first efforts to challenge models to reason with information from three documents at most. However, \citet{chen2019understanding} and \citet{min2019compositional} show that single-hop models can achieve good results in these multi-hop datasets.
Moreover, most models were also shown to degrade in adversarial evaluation~\citep{perez2020unsupervised}, where word-matching reasoning shortcuts are suppressed by extra adversarial documents~\citep{Jiang2019reasoningshortcut}.
In the \textsc{HotpotQA} \textit{open-domain} setting, the two supporting documents can be accurately retrieved by a neural model exploiting a single hyperlink~\citep{nie2019revealing,asai2019learning}.

\begin{figure*}[t]
    \centering
    \includegraphics[width=0.95\textwidth] {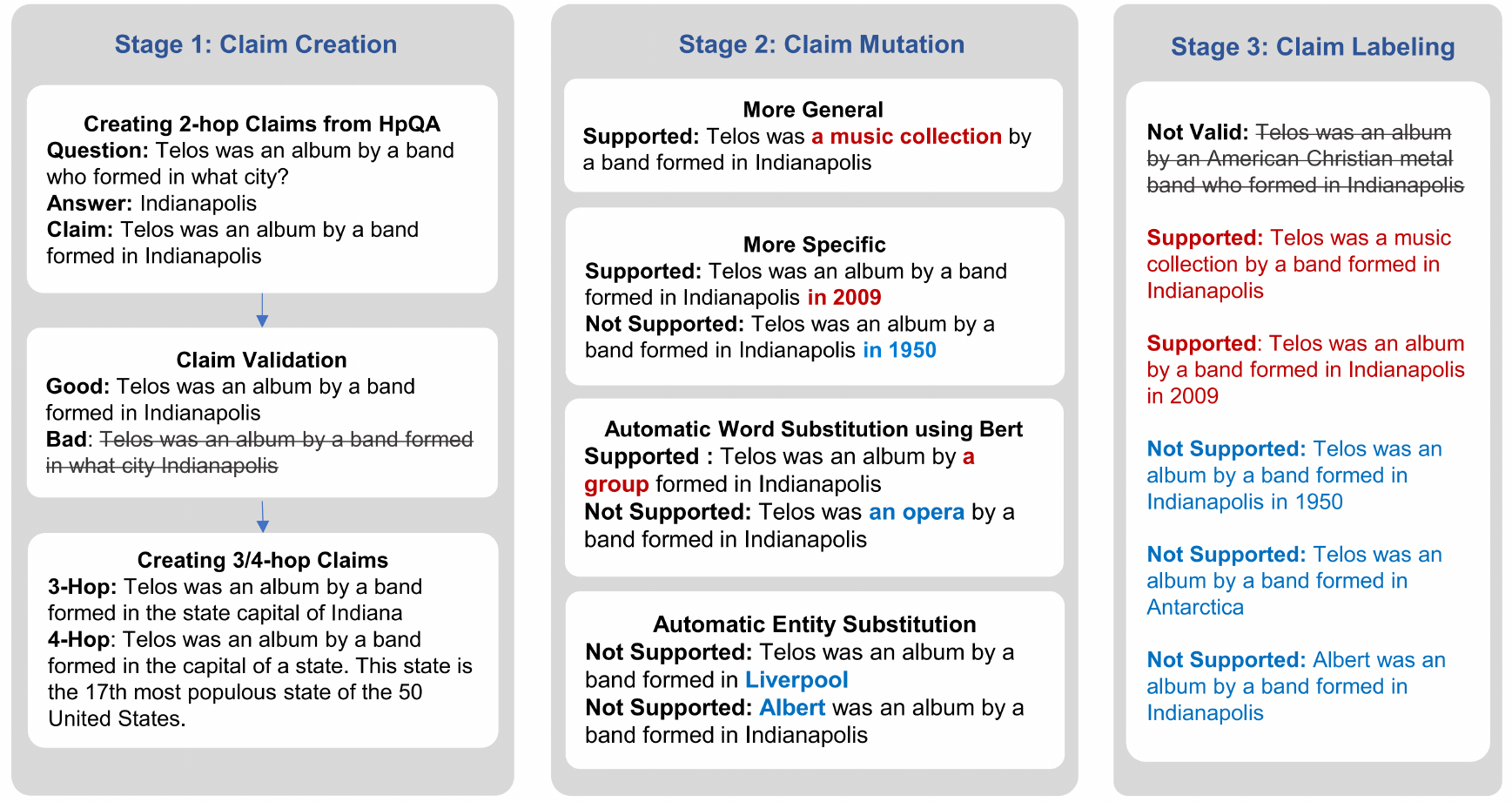}
    \vspace{-5pt}
    \caption{Data Collection flow chart for \dataset. In the first stage, we create claims from \textsc{HotpotQA}, validate them and extend to more hops. In the second stage, we apply a variety of mutations to the claims performed by crowd-workers and automatic methods. In the final stage, we ask crowd-workers to label the resulting claims.
    \vspace{-10pt}}
    \label{fig:Data_Collection}
\end{figure*}

Hence, while providing very useful starting points for the community, FEVER is mostly restricted to a single-hop setting and existing multi-hop QA datasets are limited by the number of reasoning steps and the word overlapping between the question and all evidence. An ideal multi-hop example should have at least one piece of evidence (supporting document) that cannot be retrieved with high precision by shallowly performing direct semantic matching with only the claim.
Instead, uncovering this document requires information from previously retrieved documents.
In this paper, we try to address these issues by creating \dataset{} (i.e., \textsc{Ho}ppy \textsc{Ver}ification) whose claims (1) require evidence from as many as four English Wikipedia articles and (2) contain significantly less semantic overlap between the claims and some supporting documents to avoid reasoning shortcuts.
We create \dataset{} with \datasize{} claims in three stages.
In stage 1 (left box in~\figref{fig:Data_Collection}), we ask a group of trained and evaluated crowd-workers to rewrite the question-answer pairs from \textsc{HotpotQA}~\citep{yang2018hotpotqa} into claims that mention facts from two English Wikipedia articles. 
We then introduce extra hops\footnote{The number of hops of a claim is the same as the number of supporting documents for this claim.} to a subset of these 2-hop claims by asking crowd-workers to substitute an entity in the claim with information from another English Wikipedia article that describes the original entity.
We then repeat this process on these 3-hop claims to further create 4-hop claims. 
To make many-hop claims more natural and readable, we encourage crowd-workers to write the 3/4-hop claims in multiple sentences and connect them using coreference.
An entire evolution history from 2-hop claims to 3/4-hop claims is presented in the leftmost box in \figref{fig:Data_Collection} and \tabref{table:reasoning-graphs}, where the latter further presents the reasoning graphs of various shapes embodied by the many-hop claims.

In stage 2 (the central box in~\figref{fig:Data_Collection}), we create claims that are not supported by the evidence by mutating the claims collected in stage 1 with a combination of automatic word/entity substitution and human editing. 
Specifically, we ask the trained crowd-workers to rewrite a claim by making it either more specific/general than or negating the original claim.
We ensure the quality of the machine-generated claims using human validation detailed in \secref{ssec:claim_mutation}. 
In stage 3, we follow~\citet{thorne2018fever} to label the claims as \textsc{Supported}, \textsc{Refuted}, or \textsc{NotEnoughInfo}.
However, we find that the decision between \textsc{Refuted} and \textsc{NotEnoughInfo} can be ambiguous in many-hop claims and even the high-quality, trained annotators from Appen, instead of Mturk, cannot consistently choose the correct label from these two classes.
Recent works~\citep{pavlick2019inherent,chen2020uncertain} have raised concern over the uncertainty of NLI tasks with categorical labels and proposed to shift to a probabilistic scale.
Since this work is mainly targeting the many-hop retrieval, we combine the \textsc{Refuted} and \textsc{NotEnoughInfo} into a single class, namely \textsc{Not-Supported}.
This binary classification task is still challenging for models given the incomplete evidence retrieved, as we will explain later. 

Next, we introduce the baseline system and demonstrate its limited ability in addressing many-hop claims.
Following a state-of-the-art system~\citep{nie2019combining} for FEVER, we build the baseline with a TF-IDF document retrieval stage and three BERT models fine-tuned to conduct document retrieval, sentence selection, and claim verification respectively.
We show that the bi-gram TF-IDF~\citep{chen2017reading}'s top-100 retrieved documents can only recover all supporting documents in 80\% of 2-hop claims, 39\% of 3-hop claims, and 15\% of 4-hop claims.
The performance of downstream neural document and sentence retrieval models also degrades significantly as the number of supporting documents increases. 
These results suggest that the possibility of a word-matching shortcut is reduced significantly in 3/4-hop claims.
Because the complete set of evidence cannot be retrieved for most claims, the claim verification model only achieves 73.7\% accuracy in classifying the claims as \textsc{Supported} or \textsc{Not-Supported}, while the model given all evidence predicts 81.2\% of the claims correctly under this oracle setting.
We further provide a sanity check to show that the model can only correctly predict the labels for 63.7\% of claims without any evidence. 
This suggests that the claims contain limited clues that can be exploited independently of the evidence during the verification, and a strong retrieval method capable of many-hop reasoning can improve the claim verification accuracy. 
In terms of \textsc{HoVer} as an integrated task, the best pipeline can only retrieve the complete set of evidence \textbf{and} correctly verify the claim for 14.9\% of dev set examples, falling behind the 81\% human performance significantly.

Overall, we provide the community with a novel, challenging and large many-hop fact extraction and claim verification dataset with over \datasize{} claims that can be comprised of multiple sentences connected by coreference, and require evidence from as many as four Wikipedia articles. 
We verify that the claims are challenging, especially in the 3/4-hop cases, by showing the limited performance of a state-of-the-art system for both retrieval and verification. 
We hope that the introduction of \dataset{} and the accompanying evaluation task will encourage research in complex many-hop reasoning for fact extraction and claim verification.
\section{Data Collection}
The many-hop fact verification dataset, \dataset{}, is a collection of human-written claims about facts in English Wikipedia articles created in three main stages (shown in \figref{fig:Data_Collection}).
In the \textbf{Claim Creation} stage (\secref{ssec:claim_creation}), we ask trained annotators on Appen\footnote{Previously known as Figure-Eight and CrowdFlower: \url{https://www.appen.com/}} to create claims by rewriting question-answer pairs (\secref{sssec:simple_claim_creation}) from the \textsc{HotpotQA} dataset\footnote{Because of the complexity and costs (\secref{ssec:annotators}) of the data collection pipeline, we only use the \textsc{HotpotQA} dev set and 5000 examples from the training set.} \citep{yang2018hotpotqa}.
The validated 2-hop claims are then extended to (\secref{sssec:extension}) include facts from more Wikipedia articles.
In the \textbf{Claim Mutation} stage (\secref{ssec:claim_mutation}), claims generated from the above two processes are mutated with human editing and automatic word substitution.
Finally, in the \textbf{Claim Labeling} stage (\secref{ssec:claim_labeling}), trained crowd-workers classify the original and mutated claims as either \textsc{Supported}, \textsc{Refuted} or \textsc{NotEnoughInfo}.
We merge the latter two labels into a single \textsc{Not-Supported} class, owing to ambiguity explained in \secref{ssec:claim_labeling}.
The guidelines and design for every task are shown in the appendix.

\subsection {Claim Creation}
\label{ssec:claim_creation}
The goal is to create claims by rewriting question-answer pairs from \textsc{HotpotQA} \cite{yang2018hotpotqa} and extend these claims to include facts from more documents (shown in the left box in \figref{fig:Data_Collection}).

\subsubsection{Creating 2-Hop Claims from \textsc{HotpotQA}}
\label{sssec:simple_claim_creation}
To begin with, crowd-workers are asked to combine question-answer pairs to write claims. 
These claims require information from two Wikipedia articles. 
Based on the guidelines, the annotators can neither exclude any information from the original QA pairs nor introduce any new information. 

\paragraph{Validating Created Claims.}
\label{para:claim_validation}
We then train another group of crowd-workers to validate the claims created from \secref{sssec:simple_claim_creation}. 
To ensure the quality of the claims, we only keep those where at least two out of three annotators agree that it is a valid statement and covers the same information from the original question-answer pair. 
These \textbf{validated} 2-hop claims are automatically labeled as \textsc{Supported}. 

\subsubsection{Extending to 3-Hop and 4-Hop Claims}
\label{sssec:extension}
Consider a valid 2-hop claim $\mathbf{c}$ from \secref{para:claim_validation} that includes facts from 2 supporting documents $\mathbf{A} = \{a_1,a_2\}$. 
We extend $\mathbf{c}$ to a new, 3-hop claim $\hat{\mathbf{c}}$ by substituting a named entity $\mathbf{e}$ in $\mathbf{c}$ with information from another English Wikipedia article $a_3$ that describes $\mathbf{e}$.
The resulting 3-hop claim $\hat{\mathbf{c}}$ hence has 3 supporting document $\{a_1,a_2,a_3\}$. 
We then repeat this process to extend the 3-hop claims to include facts from the forth documents.
We use two methods to substitute different entities $\mathbf{e}$, leading to 4-hop
claims with various reasoning graphs. 

\paragraph{Method 1.}
We consider the entity $\mathbf{e}$ to be the title of a document $a_k \in \mathbf{A}$.
We search for English Wikipedia articles $\hat{a} \notin \mathbf{A}$ whose text body mentions $\mathbf{e}$'s hyperlink.
We exclude the $\hat{a}$ whose title is mentioned in the text body of one of the document in $\mathbf{A}$.
We then ask crowd-workers to select $a_3$ from a candidate group of $\hat{a}$ and write the 3-hop claim $\hat{\mathbf{c}}$ by replacing $\mathbf{e}$ in $\mathbf{c}$ with a relative clause or phrase using information from a sentence $s \in a_3$.

\paragraph{Method 2.}
In this method, we consider $\mathbf{e}$ to be any other entity in the claim, which is \textbf{not} the title of a document $a_k \in \mathbf{A}$ but exists as a Wiki hyperlink in the text body of one document in $\mathbf{A}$.
The last 4-hop claim in \tabref{table:reasoning-graphs} is created via this method and the entity $\mathbf{e}$ is ``NASCAR".
The remaining efforts are the same as Method 1 as we search for English Wikipedia articles $\hat{a} \notin \mathbf{A}$ whose text body mentions $\mathbf{e}$'s hyperlink and ask crowd-workers to replace $\mathbf{e}$ with information from $a_3$.

\paragraph{Task Setup.}
We employ Method 1 to extend the collected 2-hop claims, for which we can find at least one $\hat{a}$.
Then we use both Method 1 and Method 2 to extend the 3-hop claims to 4-hop claims of various reasoning graphs.
In a 3-document reasoning graph (a chain), the title of the middle document is substituted out during the extension from the 2-hop claim and thus does not exist in the 3-hop claim.
Therefore, Method 1, which replaces the title of one of the three documents in the claim, can only be applied to either the leftmost or the rightmost document. 
In order to append the fourth document to the middle document in the 3-hop reasoning chain, we have to substitute a non-title entity in the 3-hop claim, which can be achieved by Method 2.
In \tabref{table:reasoning-graphs}, the last 4-hop claim with a star-shape reasoning graph is the result of applying Method 1 for 3-hop extension and Method 2 for the 4-hop extension, while the first two 4-hop claims are created by applying Method 1 twice.
We ask the crowd-workers to submit the index of the sentence and add this sentence to the supporting facts of the 2-hop claim to form the supporting facts of this new, 3-hop claim.

\subsection{Claim Mutation}
\label{ssec:claim_mutation}
We mutate the claims created in \secref{ssec:claim_creation} to collect new claims that are not necessarily supported by the facts.
We employ four types of mutation methods (shown in the middle column of \figref{fig:Data_Collection}) that are explained in the following sections. 

\paragraph{Making a Claim More Specific or General.} 
A \textit{more specific} claim contains information that is not in the original claim. 
A \textit{more general} claim contains less information than the original one.
We design guidelines (shown in the appendix) and quizzes to train the annotators to use natural logic. 
We constrain the annotators from replacing the supporting document titles in a claim to ensure that verifying this claim requires the same set of evidence as the original claims. 
We also forbid mutating location entities (e.g., Manhattan $\xrightarrow{}$ New York) as this may introduce external evidence (``Manhattan is in New York") that is not in the original set of evidence.

\paragraph{Automatic Word Substitution.}  

In this mutation process, we first sample a word from the claim that is neither a named entity nor a stopword.
We then use a BERT-large model~\citep{devlin2018bert} to predict this masked token, as we found that human annotators usually fall into a small, fixed vocabulary when thinking of the new word.
We ask 3 annotators to validate whether each claim mutated by BERT is logical and grammatical to further ensure the quality and keep the claims where at least 2 workers decide it suffices our criteria.
500 BERT-mutated claims passed the validation and labeling.

\paragraph{Automatic Entity Substitution.}
We design a separate mutation process to substitute named entities in the claims.
First, we perform Named Entity Recognition on the claims. 
We then randomly select a named entity that is not the title of any supporting document, and replace it with an entity of the same type sampled from the \textit{context}.\footnote{The eight distracting documents selected by TF-IDF.} 

\paragraph{Claim Negation.}
Understanding negation cues and their scope is of significant importance to NLP models. 
Hence, we ask crowd-workers to negate the claims by removing or adding negation words (e.g., \textit{not}), or substituting a phrase with its antonyms. 
However, it is shown in \citet{schuster2019towards} that models can exploit this bias as most claims containing a negation word have the label \textsc{Refuted}.
To mitigate this bias, we only include a subset of negated 2-hop claims where 60\% of them don't include any explicit negation word.

\subsection{Claim Labeling}
\label{ssec:claim_labeling}
In this stage (the right column in \figref{fig:Data_Collection}), we ask annotators to assign one of the three labels (\textsc{Supported}, \textsc{Refuted}, or \textsc{NotEnoughInfo}) to all 3/4-hop claims (original and mutated) as well as 2-hop mutated claims. 
The workers are asked to make judgments based on the given supporting facts solely without using any external knowledge. 
Each claim is annotated by five crowd-workers and we only keep those claims where at least three agree on the same label, resulting in a fleiss-kappa inter-annotator agreement score of 0.63.\footnote{We discarded a total of 2222 claims that received a vote of 2 vs 2 vs 1. They only account for less than 10\% of all the claims that we have kept in the dataset.}

\paragraph{\textsc{Not-Supported} Claims.}
\label{para:nonsupp}
The demarcation between \textsc{NotEnoughInfo} or \textsc{Refuted} is subjective and the threshold could vary based on the world knowledge and perspective of annotators. Consider the claim \textit{``Christian Bale starred in a 2010 movie directed by an American director"} and the fact \textit{``English director Christopher Nolan directed the Dark Knight in 2010"}.
Although the ``\textit{American}" in the claim directly contradicts the word ``\textit{English}" in the fact, this claim should still be classified as \textsc{NotEnoughInfo} as Bale could have starred in another 2010 film by an American director.
More of such examples are provided in the appendix.
In this case, a piece of evidence contradicts a relative clause in the claim but does not refute the entire claim. 
Similar problems regarding the uncertainty of NLI tasks have been pointed out in previous works~\cite{zaenen2005local,pavlick2019inherent,chen2020uncertain}.

We design an exhaustive list of rules with abundant examples, trying to standardize the decision process for the labeling task. 
We acknowledge the difficulty and cognitive load it sometimes bears on well-informed annotators to think of corner cases like the example shown above. 
The final annotated data revealed the ambiguity between \textsc{NotEnoughInfo} and \textsc{Refuted} labels, as in a 100-sample human validation, only 63\% of the labels assigned by another annotator match the majority labels collected.
Hence we combine the \textsc{Refuted} and \textsc{NotEnoughInfo} into a single class, namely \textsc{Not-Supported}.
90\% of the validation labels match the annotated labels under this binary classification setting.

\subsection{Annotator Details}
\label{ssec:annotators}
Most annotators are native English speakers from the UK, US, and Canada. 
For all tasks, we first launch small-scale pilots to train annotators and incorporate their feedback for at least two rounds. 
Then for claim creation and extension tasks, we manually evaluate the claims they created and only keep those workers who can write claims of high quality.
For claim validation (\secref{sssec:simple_claim_creation}) and labeling (\secref{ssec:claim_labeling}) tasks, we additionally launch quizzes and annotators scoring 80\% accuracy in the quiz are then admitted to the job. 
During the job, we use test questions to ensure their consistent performance. 
Crowd-workers whose test-question accuracy drops below 82\% are rejected from the tasks and all his/her annotations are re-annotated by other qualified workers.
As suggested in \citet{ramirez2019understanding}, we highlight the mutated words during the labeling tasks to reduce the mental workload on workers and speed up the jobs. 
The crowd-workers are paid an average of 12 cents (the pay varies with the number of hops of a claim) per hit; and for the hop extension job, they are paid as much as 40 cents per hit since the task is time-consuming and demands the annotators to rewrite the claims after incorporating information from the extra document.

\begin{table}[t]
\centering
\begin{small}
\begin{tabular}[t]{cc|cc|c}
\toprule
Split & \#Hops & \textsc{Supported} & \textsc{Not-Sup} & \textsc{Total} \\
\midrule
\multirow{4}{*}{Train}
& 2 & 6496 & 2556 & 9052 \\
& 3 & 3271 & 2813 & 6084\\
& 4 & 1256 & 1779 & 3035\\
& Total & 11023 & 7148 & 18171\\
\midrule
\multirow{4}{*}{Dev} & 2 & 521 & 605 & 1126 \\
& 3 & 968 & 867 & 1835 \\
& 4 & 511 & 528 & 1039 \\
& Total & 2000 & 2000 & 4000\\
\midrule
Test & - & 2000 & 2000 & 4000\\
\midrule
Total & - & 15023 & 11148 & 26171 \\
\bottomrule
\end{tabular}
\vspace{-5pt}
\caption{The sizes of the Train-Dev-Test split for \textsc{Supported} and \textsc{Not-Supported} classes and different number of hops.
 \vspace{-10pt}
}
\label{table:data_stat}
\end{small}
\end{table}
 
\section{Dataset Analysis}

\paragraph{Dataset Statistics.}
We partitioned the annotated claims and evidence into training, development, and test sets.
The detailed statistics are shown in \tabref{table:data_stat}.
Because of the job complexity, judgment time, and the difficulty of quality control (described in \secref{ssec:annotators}) increase drastically along with the number of hops of the claim, the first version of \dataset{} only uses 12k examples from the \textsc{HotpotQA}~\citep{yang2018hotpotqa}.
The 2-hop, 3-hop and 4-hop claims have a mean length of 19.0, 24.2, and 31.6 tokens respectively as compared to a mean length of 9.4 tokens in \citet{thorne2018fever}.

\paragraph{Diverse Many-Hop Reasoning Graphs.}
As questions from \textsc{HotpotQA}~\citep{yang2018hotpotqa} require two supporting documents, our 2-hop claims created from \textsc{HotpotQA} question-answer pairs inherit the same 2-node reasoning graph as shown in the first row in \tabref{table:reasoning-graphs}.
However, as we extend the original 2-hop claims to more hops using approaches described in \secref{sssec:extension}, we achieve many-hop claims with diverse reasoning graphs. 
Every node in a reasoning graph is a unique document that contains evidence, and an edge that connects two nodes represents a hyperlink from the original Wikipedia document or a comparison between two titles.
As shown in \tabref{table:reasoning-graphs}, we have three unique 4-hop reasoning graphs that are derived from the 3-hop reasoning graph by appending the 4th node to one of the existing nodes in the graph. 

\paragraph{Qualitative Analysis.} 
The process of removing a bridge entity and replacing it with a relative clause or phrase adds a lot of information to a single hypothesis.  
Therefore, some of the 3/4-hop claims are of relatively longer length and have complex syntactic and reasoning structure. In systematic aptitude tests as well, humans are assessed on synthetically designed complex logical puzzles. These tests require critical problem solving abilities and are effective in evaluating logical reasoning capabilities of humans and AI models.
Overly complicated claims are discarded in our labeling stage if they are reported as ungrammatical or incomprehensible by the annotators. The resulting examples form a challenging task of evidence retrieval and multi-hop reasoning.
\section{Baseline System}
Following a state-of-the-art system~\cite{nie2019combining} on FEVER~\citep{thorne2018fever}, we build a pipeline system of fact extraction and claim verification.\footnote{We provide a simple visualization of the entire pipeline in the appendix. }
This provides an initial baseline for future works and its performance indicates the many-hop challenge posed by \dataset{}.

\paragraph{Rule-based Document Retrieval.}
We use the document retrieval component from~\citet{chen2017reading} that returns the k closest Wikipedia documents for a query using cosine similarity between binned uni-gram and bi-gram TF-IDF vectors.
This step outputs a set $\mathbf{P_r}$ of $k_r$ document that are processed by downstream neural models.

\paragraph{Neural-based Document Retrieval.}
Similar to the retrieval model in~\citet{nie2019combining}, the BERT-base model~\cite{devlin2018bert} takes a single document $p \in \mathbf{P_r}$ and the claim $c$ as the input, and outputs a score that reflects the relatedness between $p$ and $c$.
We select a set $\mathbf{P_n}$ of top $k_p$ documents having relatedness scores higher than a threshold of $\kappa_p$.

\paragraph{Neural-based Sentence Selection.}
We fine-tune another BERT-base model that encodes the claim $c$ and all sentences from a single document $p \in \mathbf{P_n}$, and predicts the sentence relatedness scores using the first token of every sentence.
We select a set $\mathbf{S_n}$ of top sentences from the entire $\mathbf{P_n}$ having relatedness scores higher than a threshold of $\kappa_s$.

\paragraph{Claim Verification Model.}
We fine-tune a BERT-base model for recognizing textual entailment between the claim $c$ and the retrieved evidence $\mathbf{S_n}$.
We feed the claim and retrieved evidence, separated by a [SEP] token, as the input to the model and perform a binary classification based on the output representation of the [CLS] token at the first position.
\section{Experiments and Results}
We explain the evaluation metrics we use and report the results of the baseline in three evaluation tasks.

\begin{table}[t]
\centering
\begin{small}
\begin{tabular}[t]{l|ccc|c}
\toprule
& \multicolumn{3}{c}{\textbf{\#Hops}} & \multirow{2}{*}{\textbf{Overall}}\\
Hit@ & 2 & 3 & 4 & \\
\midrule
5 & 42.10 & 9.97 & 0.38& 16.53\\
10 & 53.37 & 15.91 & 2.89 & 23.08 \\
25 & 66.16 & 24.90 & 6.83 & 31.83 \\
100 & 80.02 & 39.18 & 15.59 & 44.55 \\
\bottomrule
\end{tabular}
\vspace{-5pt}
\caption{The performance of the TF-IDF Document Retrieval, evaluated on the supported claims in the dev set.  
 \vspace{-5pt}
}
\label{table:rule_doc_ret}
\end{small}
\end{table}

\begin{table}[t]
\centering
\begin{small}
\resizebox{0.48\textwidth}{!}{%
\begin{tabular}[t]{l|ccc|c}
\toprule
& \multicolumn{3}{c}{\textbf{\#Hops}} & \multirow{2}{*}{\textbf{Overall}}\\
\textbf{Models} & 2 & 3 & 4 & \\
\midrule
BERT & 30.1/69.5 & 5.6/57.6 & 0.6/52.6 & 11.2/59.1 \\
BERT$^{\star}$ & 34.0/69.9 & 5.8/58.2 & 1.0/53.4 & 12.5/60.2  \\
\midrule
Oracle & 50.9/81.7 & 28.1/79.1 & 26.2/82.2 & 34.0/80.6 \\
Human & 85.0/92.5 & 82.4/95.3 & 65.8/91.4 & 77.0/93.5 \\
\bottomrule
\end{tabular}}
\vspace{-5pt}
\caption{The EM/F1 scores of the document retrieval methods, evaluated on the dev set.
 \vspace{-5pt}
}
\label{table:neural_doc_ret}
\end{small}
\end{table}

\begin{table}[t]
\centering
\begin{small}
\resizebox{0.48\textwidth}{!}{%
\begin{tabular}[t]{l|ccc|c}
\toprule
& \multicolumn{3}{c}{\textbf{\#Hops}} & \multirow{2}{*}{\textbf{Overall}}\\
\textbf{Models} & 2 & 3 & 4 & \\
\midrule
BERT & 13.6/57.2 & 1.9/49.8 & 0.2/45.0 & 4.8/50.6 \\
BERT$^{\star}$ & 9.1/52.0 & 1.3/45.4 & 0.3/41.2 & 3.2/46.2 \\
\midrule
Oracle & 25.0/68.3 & 18.4/71.5 & 17.1/76.4 & 19.9/71.9 \\
Human & 75.0/86.5 & 73.5/93.1 & 42.1/87.3 & 56.0/88.7 \\
\bottomrule
\end{tabular}}
\vspace{-5pt}
\caption{The EM/F1 scores of the sentence retrieval methods, evaluated on the dev set.
 \vspace{-5pt}
}
\label{table:neural_sent_ret}
\end{small}
\end{table}

\begin{table}[t]
\centering
\begin{small}
\begin{tabular}[t]{l|ccc|c}
\toprule
& \multicolumn{3}{c}{\textbf{\#Hops}} & \multirow{2}{*}{\textbf{Overall}}\\
\textbf{Models} & 2 & 3 & 4 &\\
\midrule
BERT + \textsc{Oracle} & 79.8 & 83.5 & 78.6 & 81.2 \\
Claim-only & 57.5 & 67.7 & 63.6 & 63.7 \\
Human + \textsc{Oracle} & 92.6 & 88.4 & 87.2 & 90.0 \\
\bottomrule
\end{tabular}
\vspace{-5pt}
\caption{The claim verification accuracy of the NLI models, evaluated on the dev set.
 \vspace{-5pt}
}
\label{table:NLI}
\end{small}
\end{table}

\begin{table}[t]
\centering
\begin{small}
\begin{tabular}[t]{l|cc}
\toprule
\textbf{Models}& \textbf{Accuracy(\%)} & \textbf{\textsc{HoVer} Score (\%)}\\
\midrule
BERT + \textsc{Gold} & 67.6 & 14.9 \\
BERT + \textsc{Retr} & 73.7 & 14.5 \\
Human & 88.0 & 81.0\\
\bottomrule
\end{tabular}
\vspace{-5pt}
\caption{The claim verification accuracy and \textsc{HoVer} scores of the entire pipeline, evaluated on the dev set.
 \vspace{-5pt}
}
\label{table:full-pip}
\end{small}
\end{table}

\begin{table}[t]
\centering
\begin{small}
\begin{tabular}[t]{l|cc}
\toprule
\textbf{Model} & \textbf{Evidence F1} & \textbf{\textsc{HoVer} Score (\%)} \\
\midrule
BERT & 49.5 & 15.32 \\
\bottomrule
\end{tabular}
\vspace{-5pt}
\caption{The evidence F1 score and \textsc{HoVer} score of the best model, evaluated on the test set.
 \vspace{-5pt}
}
\label{table:full-pip-test}
\end{small}
\end{table}

\subsection{Evaluation Metrics}
We evaluate the final accuracy of the claim verification task to predict a claim as \textsc{Supported} or \textsc{Not-Supported}.
The document and sentence retrieval are evaluated by the exact-match and F1 scores between the predicted document/sentence-level evidence and the ground-truth evidence for the claim. We refer to the appendix for the detailed experimental setups and hyper-parameters.

\subsection{Document Retrieval Results}
\label{ssec:doc_retrieval_results}
The results in \tabref{table:rule_doc_ret} show the task becomes significantly harder for the bi-gram TF-IDF when the number of supporting documents increases.
This decline in single-hop word-matching retrieval rate suggests that the method to extend the    reasoning hops (\secref{sssec:extension}) is effective in terms of promoting multi-hop document retrieval and minimizing word-matching reasoning shortcuts.
We then use a BERT-base model (1st row in \tabref{table:neural_doc_ret}) to re-rank the top-20 documents returned by the TF-IDF.
The ``BERT$^{\star}$" (2nd row) is trained with an oracle training set containing all golden documents.
Overall, the performances of the neural models are limited by the low recall of the 20 input documents and the F1 scores degrade as the number of hops increase.
The oracle model (3rd row) is the same as ``BERT$^{\star}$" but evaluated on the oracle data.
It indicates an upper bound of the BERT retrieval model given a perfect rule-based retrieval method.
These findings again demonstrate the high quality of the many-hop claims we collected, for which the reasoning shortcuts are significantly reduced because of the approach described in \secref{sssec:extension}.

\subsection{Sentence Selection Results}
\label{ssec:sentence_selection_results}
We evaluate the neural-based sentence selection models by re-ranking the sentences within the top-5 documents returned by the best neural document retrieval method.
For ``BERT$^{\star}$" (2nd row in \tabref{table:neural_sent_ret}), we again ensured that all golden documents are contained within the 5 input documents during the training.
We then measure the oracle result by evaluating ``BERT$^{\star}$" on the dev set with all golden documents presented.
This suggests an upper bound of the sentence retrieval model given a perfect document retrieval method.
The same trend holds as the F1 scores decrease significantly as the number of hops increases.\footnote{The only exception is in the oracle setting because selecting sentences from 4 out of 5 documents is actually easier than selecting from 2 out of 5 documents.}

\subsection{Claim Verification Results}
\label{ssec:claim_verification__results}
In an oracle (1st row in \tabref{table:NLI}) setting where the complete set of evidence is provided, the model achieves 81.2\% accuracy in verifying the claims.
We also conduct a sanity check in a claim-only environment (2nd row) where the model can only exploit the bias in the claims without any evidence, in which the model achieves 63.7\% accuracy.
Although the model can exploit limited biases within the claims to achieve higher-than-random accuracy without any evidence, it is still 17.5\% worse than the model given the complete evidence.
This suggests the NLI model can benefit from an accurate evidence retrieval model significantly.

\subsection{Full Pipeline Results}
The full pipeline (``BERT+Retr" in \tabref{table:full-pip}) uses the sentence-level evidence retrieved by the best document/sentence retrieval models as the input to the NLI model, while the ``BERT+Gold" is the oracle in \tabref{table:NLI} but evaluated with retrieved evidence instead.
We further propose the \textsc{HoVer} Score, which is the percentage of the examples where the the model must retrieve at least one supporting fact from every supporting document and predict the correct label.
We show the performance of the best model (BERT+Gold in \tabref{table:full-pip}) on the test set in \tabref{table:full-pip-test}.
Overall, the best pipeline can only retrieve the complete set of evidence and predict the correct label for 14.9\% of examples on the dev set and 15.32\% of examples on the test set, suggesting that our task is indeed more challenging than the previous work of this kind.

\subsection{Human Performance}
We measure the human performance on 100 sampled claims.
In the document (\tabref{table:neural_doc_ret}) and sentence retrieval (\tabref{table:neural_sent_ret}) tasks, the human F1 score is 37.9\% and 33.1\% higher than the best baseline respectively.
In the oracle claim verification (\tabref{table:NLI}), the human accuracy is 90\%, i.e., 8.8\% higher than BERT's accuracy. 
Comparing on the full pipeline (\tabref{table:full-pip}), the human accuracy and human \textsc{HoVer} score are 88\% and 81\%, while the best BERT model only obtains 67.6\% accuracy and 14.9\% \textsc{HoVer} score respectively on the dev set.
Human evaluation setup is explained in appendix.
\section{Related Work}

\paragraph{Natural Language Inference and Fact Verification.}
Textual Entailment and natural language inference (NLI) datasets like RTE
\cite{dagan_dolan_magnini_roth_2009},  SNLI \cite{bowman2015large} or  MNLI \cite{N18-1101} consist of single sentence premise. 
In this task, every premise-hypothesis pair is labeled as \textsc{Entailment}, \textsc{Contradiction}, or \textsc{Neutral}.
Another related task is fact verification, where claims (hypothesis) are checked against facts (premise).
\citet{vlachos-riedel-2014-fact} and \citet{ferreira2016emergent}
collected statements from PolitiFact, a Pulitzer Prize-winning fact-checking website that covers political topics. 
The veracity of these facts is crowd-sourced from journalists, public figures and ordinary citizens. 
However, developing  machine learning based assessments on datasets with less than five hundred datapoints is not feasible. 
\citet{wang2017liar} introduced \textsc{Liar} which includes 12,832 labeled claims from PolitiFact. The dataset is based on the metadata of the speaker and their judgments. However, the evidence supporting the statements are not provided. 
A recent work in Table-based fact verification~\citep{chen2020tablefact} points out the difficulty of collecting accurate neutral labels and leaves out those neutral claims at the claim creation phase.
We instead merge neutral (\textsc{NotEnoughInfo}) claims with \textsc{Refuted} claims into a single class.

\paragraph{Fact Extraction and Verification.}  
\citet{thorne2018fever} introduced FEVER, a fact extraction and verification dataset. It consists of single sentence claims that are verified against the pieces of evidence retrieved from at most two documents. In our dataset, the claims vary in size from one sentence to one paragraph and the pieces of evidence are derived from information ranging from one document to four documents. More recently, \citet{thorne2019fever2} introduced the FEVER2.0 shared task which challenge participants to fact verify claims using evidence from Wikipedia and to attack other participant's system with adversarial models.  
In \dataset{}, the claim needs verification from multiple documents. 
Prior to verification, the relevant documents and the context inside these documents must also be retrieved accurately.
More recently, \citet{chen2019seeing} enriched the claim with multiple perspectives that support or oppose the claim in different scale. 
Each perspective can also be verified by existing facts. 
MultiFC~\cite{augenstein2019multifc} is a dataset of naturally occurred claims from multiple domains.
The contribution of these two fact-checking dataset is orthogonal to ours.

\paragraph{Multi-Hop Reasoning Datasets.}
Many recently proposed datasets are created to challenge models' ability to reason across multiple sentences or documents.
\citet{khashabi2018looking} introduced Multi-Sentence Reading Comprehension (MultiRC) which is composed of 6k multi-sentence questions. 
\citet{Mihaylov2018CanAS} introduced Open Book Question Answering composed of 6000 questions created upon 1326 science facts.
It requires combining an open book fact with broad common knowledge in a multi-hop reasoning process. 
\citet{welbl2018constructing} constructed a multi-hop QA dataset, QAngaroo, whose queries are automatically generated upon an external knowledge base.
\citet{yang2018hotpotqa} introduced the \textsc{HotpotQA} dataset which does not rely on an external knowledge base and provides sentence-level evidence to explain the answer. 
Recent state of the art models on the open-domain setting of \textsc{HotpotQA} include \citet{nie2019revealing,qi2019answering,asai2019learning,fang2019hierarchical,Zhao2020Transformer-XH}. The dataset is diverse and natural as it is created by human annotators.
ComplexWebQuestion~\cite{talmor-berant-2018-web} built multi-hop questions by combining simple questions (paired with SPARQL queries) followed by human paraphrasing. 
Thus, each question is annotated with not only the answer, but also the single-hop queries that can be used as the intermediate supervision.
\citet{jansen-2018-worldtreev1} pointed out the difficulty of information aggregation when answering multi-hop questions. 
\citet{jansen-etal-2018-worldtree} and \citet{xie-etal-2020-worldtreev2} further constructed the WorldTree datasets composed of science exam questions with annotated explanations, where each question is annotated with an average of 6 supporting facts (and as many as 16 facts).
These datasets are mostly presented in the question answering format, while \dataset{} is instead created for the task of claim verification.
\citet{hidey-etal-2020-deseption} created an adversarial fact-checking dataset containing 417 composite claims that consist of multiple propositions to attack FEVER models. 
Compared to these previous efforts, \dataset{} is significantly larger in the size while also expanding the richness in language and reasoning paradigms.

\paragraph{Synthetic Datasets.} 
Winograd Schema Challenge \citep{sakaguchi2019winogrande}, Winogender schema\citep{rudinger-EtAl:2018:N18}, and RuleTaker \citep{Clark2020TransformersAS} are synthetic datasets created to challenge models' ability to understand the complex reasoning in natural language.
With the same motive, \textsc{HoVer} is created by humans following the guidelines and rules designed to enforce a multi-hop structure within the claim.
Compared to synthetic datasets like RuleTaker, \textsc{HoVer}'s examples are more natural as they are created and verified by humans and cover a wider range of vocabulary and linguistic variations. This is extremely important because models usually get close-to-perfect performance (e.g., 99\% in RuleTaker) on these synthetic datasets.
\section{Conclusion}
We present \dataset{}, a fact extraction and verification dataset requiring evidence retrieval from as many as four Wikipedia articles that form reasoning graphs of diverse shapes. 
We show that the performance of existing state-of-the-art models degrades significantly on our dataset as the number of reasoning hops increases, hence demonstrating the necessity of robust many-hop reasoning in achieving strong results.
We hope that \dataset{} will encourage the development of models capable of performing complex many-hop reasoning in the tasks of information retrieval and verification.

\section*{Acknowledgments}
\vspace{-5pt}
We thank the reviewers for their helpful comments and the annotators for their time and effort. This work was supported by DARPA MCS Grant N66001-19-2-4031, DARPA KAIROS Grant FA8750-19-2-1004, and Verisk Analytics, Inc. The views are those of the authors and not of the funding agency.

\bibliography{emnlp2020}
\bibliographystyle{acl_natbib}

\appendix
\section*{Appendix}

\begin{figure}[t]
    \centering
    \includegraphics[scale=.42]{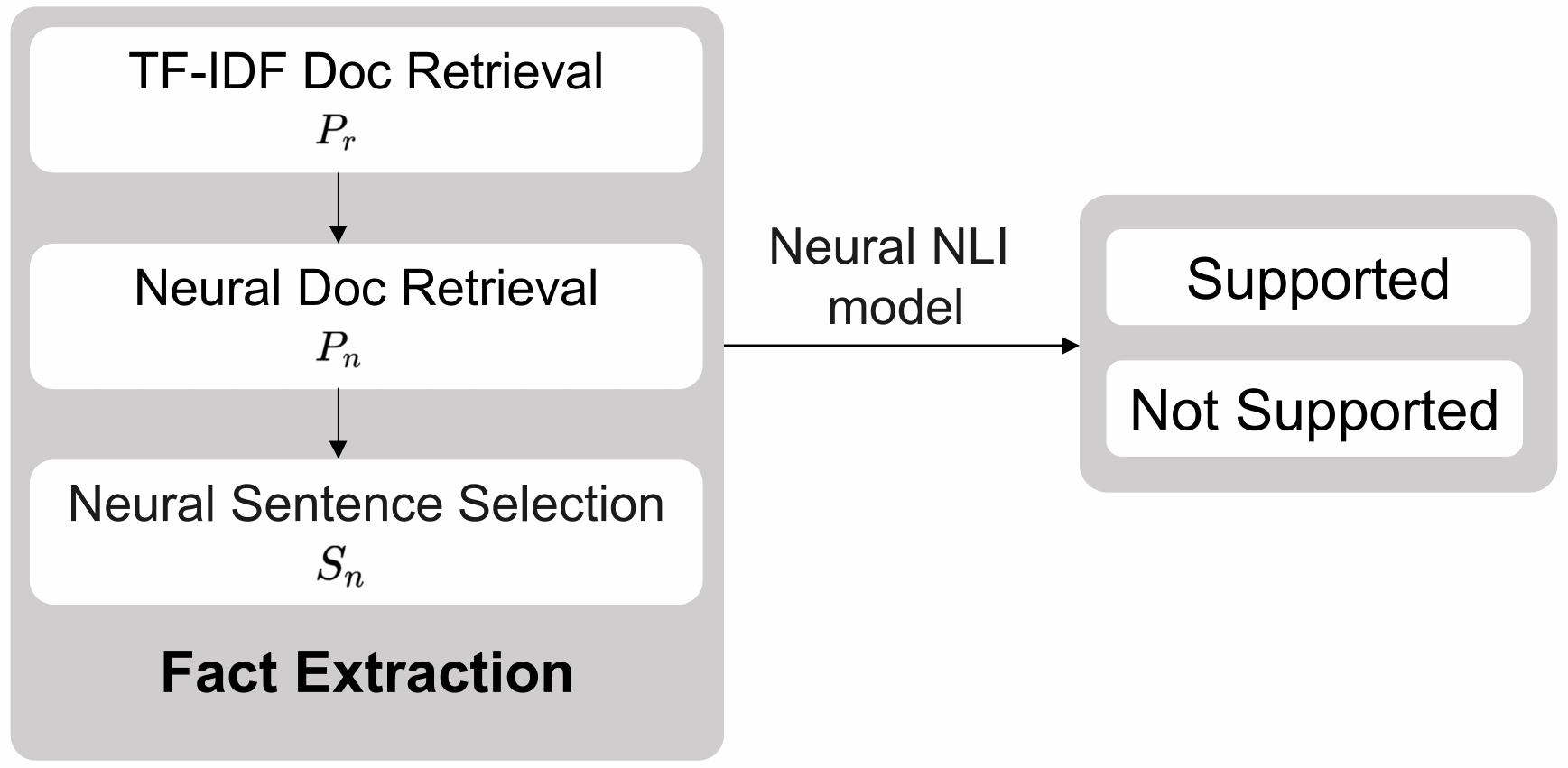}
    \caption{Baseline system with the 4-stage architecture.}
    \label{fig:Model}
\end{figure}

\section{Experimental Setup}
We use the pre-trained BERT-base uncased model (with 110M parameters) for the tasks of neural document retrieval, sentence selection, and claim verification.
The fine-tuning is done with a batch size of 16 and the default learning rate of 5e-5 without warmup.
We set $k_r = 20$, $k_p=5$, $\kappa_p=0.5$, and $\kappa_s=0.3$ based on the memory limit and the dev set performance.
We select our system with the best dev-set verification accuracy and report its scores on the hidden test set.
The entire pipeline is visualized in \figref{fig:Model}.
For document retrieval and sentence selection tasks, we fine-tune the BERT on 4 Nvidia V100 GPUs for 3 epochs. 
The training of both tasks takes around 1 hour.
For claim verification task, we fine-tune the BERT on a single Nvidia V100 for 3 epochs.
The training finishes in 30 minutes.

\paragraph{Human Evaluation}
We measure the human performance in all three evaluation tasks on 100 sampled claims.
To perform the open-domain document retrieval task, the testee is given a claim and a python program that can retrieve the Wikipedia document from the database by its title.
The testee is additionally allowed to search in the official Wikipedia web page as retrieving some documents requires matching the claim against the document content.
To select the sentence-level evidence from the retrieved documents, the testee uses the documents, tokenized by sentence, returned from the python program.
To verify the claim in the oracle setting, the testee is given all golden supporting documents.
The testee is given infinite amounts of time for each example. 
Only 2 out of 100 claims are labeled as not grammatical/logical during the human evaluation.
\begin{figure}[t]
\centering
\begin{tikzpicture}[scale=0.6, transform shape] 

\begin{axis}[ybar interval, ymax=40,ymin=0,  
     xlabel={Number of Hops},
    ylabel={Token Length},
    xmax = 5, xmin=2,]
\addplot 
coordinates { (2, 19.0)(3, 24.2)(4, 31.6)(5,30)};
\end{axis}

\end{tikzpicture}
\caption{The average token length of our 2, 3, 4-hop claims.}
\label{fig:hist}
\end{figure}
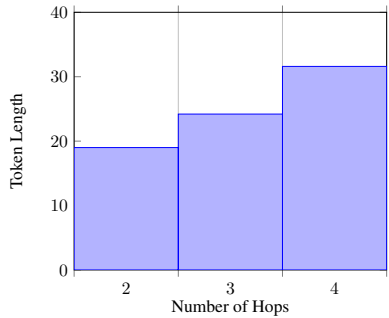

\section{Annotation Guidelines}
\label{appdx:guidelines}

\subsection{Claim Creation Guidelines}
\label{s:cc}
\paragraph{Claim.} A claim is written in single or multiple sentences that has information (true or mutated) about single or multiple entities.

\subsubsection{Simple Claim Creation}
\label{ss:scc}
The objective of this task is to generate single-sentence claims using QA pairs from \textsc{HotpotQA} dataset as shown in \figref{fig:hotpot}

\paragraph{Instructions}
\begin{itemize}
    \item Given the question and answer pair , rate the clarity of the question on a scale of 1 (very confusing) to 3 (very clear)
	\item Extract as much information as possible from the Question and Answer and rewrite them as sentences to create claims.
	\item Avoid including any extra information or uncommon words that are not part of the original Question and Answer 
	\item Claims must not exclude any information or uncommon words from the original Question and Answer
	\item Claims must not include any information beyond the question and answer
	\item Claims should be grammatically correct and in formal English
	\item Correct capitalization and spelling of entities should be followed
	\item Claims must not contain speculative language (e.g. probably, might be, maybe, etc.)
	\item Some claims might not be true
	\item Claim should be a single-sentence statement and must not contain a question mark

\end{itemize}

\subsubsection{Claim Validation}
The objective of this task is to validate whether the generated claims from \textbf{Simple Claim Creation} meet the requirements

\paragraph{Instructions}
\begin{itemize}
\item Indicate whether the  claim meets the criteria mentioned in Section \secref{ss:scc}
\item Rate the clarity of question answer pair on a scale of 1 to 5
\end{itemize}

We collect three judgments per claim and keep those claims where at least two annotators decide that it is validated.

\begin{figure}[!t]
\MakeFramed{\advance\hsize-7\width\everypar{\parshape 1 -.1cm 7cm}\FrameRestore}

\textbf{Question:} Telos was an album by a band who formed in what city?

\textbf{Answer:} Indianapolis

\textbf{Claim Created:}
Telos was an album by a band formed in Indianapolis
\endMakeFramed
   
\vspace{-0.1in}

\caption{A 2-hop Simple Claim Creation example using \textsc{HotpotQA} pair.}
\label{fig:hotpot}  
\end{figure}
    
\subsubsection{Extending to 3-hop and 4-hop}
The objective of this task is to substitute an entity in the claim with the information provided in the given English Wikipedia article.

\paragraph{Overview}

\begin{itemize}
    \item{Review the original claim and the given entity}
    \item{Select a paragraph from 1 to 5 candidate paragraphs (Every paragraph mentions the entity at least once)}
    \item{ Replace the entity with the information from your selected paragraph that describes the entity and rewrite the claim }
\end{itemize}

\paragraph{Instructions}
\begin{itemize}

    \item{The rewritten claim must contain the title of the selected paragraph (unless the title contains the entity to be replaced.)}
    
    \item{Do not fact check the information or use any external knowledge for this task}
    
    \item{The claim should be broken into multiple sentences to form a coherent paragraph }
    \item {In order to write coherent sentences, use proper pronoun/coreference in the latter sentence to properly refer to the entities mentioned in previous sentences}
    \item {The claim must not contain the entity that need to be replaced}
    \item {The claim should preserve other information from the original claim except for the entity to be replaced}
    \item {Write concise claims. Use the shortest chunk of words from one selected sentence to accurately describe the entity to be replaced}
    \item {When necessary, rephrase the claim to make it fluent and grammatically correct}

\end{itemize}

\paragraph{Example of hop-extended claims } 

\paragraph{2-hop: }
Skagen Painter Peder Severin Kroyer favored naturalism along with Theodor Esbern Philipsen and \textit{Kristian Zahrtmann.}

\paragraph{3-hop: }
Skagen Painter Peder Severin Kroyer favored naturalism along with Theodor Esbern Philipsen and \textit{the artist Ossian Elgstrom studied with in 1907.}

\subsection{Claim Mutation}

\subsubsection{Automatic Word Substitution using BERT}

\begin{figure*}[!t]
\MakeFramed{\advance\hsize-7\width\everypar{\parshape 1 -.1cm 15cm}\FrameRestore}

\textbf{Original Claim:} This Maroon 5 song, is one of the \textcolor{blue}{ \textbf{songs}} that Zaedan is best known for remixing. He is a Swedish  \textcolor{red}{\textbf{songwriter}} who worked with Taylor Swift.

\textbf{Choices:} [\textit{song, one, songs, best, known, remixing, songwriter, worked}] 

\textbf{Random Picks:} [\textit{songs, songwriter}]

\textbf{BERT Mutated Claim:} This Maroon 5 song, is one of the \textcolor{blue}{\textbf{tracks}} that Zaedan is best known for remixing. He is a Swedish \textcolor{red}{\textbf{producer}} who worked with Taylor Swift.
\endMakeFramed
\caption{Bert Mutation Procedure. We first randomly select 1-2 non-entity words from a range of \textbf{Choices} and mask them. Then the BERT model predict the masked token and provides the mutated claim.
}
\label{figure:bert_mutation}
\end{figure*}

In this mutation process, we first sample a word from the claim that is not a named entity nor a stopword. 
We then use a pre-trained BERT-large model~\citep{devlin2018bert} to predict this masked token.
We only keep the claims where (1) the new word predicted by BERT and the masked word do not have a common lemma and where (2) the cosine similarity of the BERT encoding between the masked word and the predicted word lie between $0.7$ and $0.8$.
The entire procedure is visualized in \figref{figure:bert_mutation}.

\subsubsection{Claim Negation}
\paragraph{Instructions} 
\begin{itemize}
    \item Negate the original claim even if it is inaccurate
    \item Negated claim must not include any extra information or uncommon words that are not part of the original claim
    \item Negated claim MUST include all key words, have no question mark, and must end in a period
    \item Negated claim should match the capitalization and spelling of the original claim
    \item Negated claim should not include extra information that is not part of the original claim

\end{itemize}

\subsubsection*{Examples of Negated Claims}
\paragraph{Original: }The scientific name of the \textit{true} creature featured in ``Creature from the Black Lagoon" is Eucritta melanolimnetes.

\paragraph{Negated: }The scientific name of the \textit{imaginary} creature featured in ``Creature from the Black Lagoon" is Eucritta melanolimnetes.

\subsubsection{Specifically Implied Claims}
The objective of this task is to create specifically implied claims from the  claims created in \secref{s:cc} such that the mutated claim implies the original claim.

\subsubsection{Instructions} 

\begin{itemize}
    \item  Make the claim more specific by adding information about \textit{target entities} so that the mutated claim implies the original claim.
    \item  Information must be added that is directly related to the target entities.
    \item Annotators are discouraged to verify the added information from Wikipedia or other external sources.
    \item Target entity must not be added to the mutated claim if it was not originally in the claim as it would decrease the number of hops in a claim.
    \item An entity name that is explained in a relative clause or phrase in the original claim must not be added as it would decrease the number of hops in a claim.
    
\end{itemize}

\subsubsection*{Examples of specifically implied claims}

\paragraph{Claim:}

Skagen Painter Peder Severin Kroyer favored naturalism along with Theodor Esbern Philipsen and the artist Ossian Elgstrom studied with in \textit{1907}.

\paragraph{Specifically Implied Claim:}
Skagen Painter Peder Severin Kroyer favored naturalism along with Theodor Esbern Philipsen and the \textit{muralist} Ossian Elgström studied with in 1907.

\subsubsection{Generally Implied Claims}
The objective of this task is to create generally implied claims from the  claims created in \secref{s:cc} such that the original claim implies the mutated claim.

\paragraph{Instructions} 
\begin{itemize}
    \item  Make the claim more general by deleting information about \textit{target entities} so that the original claim implies the mutated claim.
    \item Pick an entity and consider the less specific/more generic term
    \item if defender then swap for player; if goalie then player; if 1963, then 1960’s … etc.
    \item Removing information - Never remove the entire clause
    
\end{itemize}

\subsubsection*{Examples of generally implied claims}

\paragraph{Claim:}Skagen Painter Peder Severin Kroyer favored naturalism along with Theodor Esbern Philipsen and the artist Ossian Elgstrom studied with in \textit{1907}.

\paragraph{Generally Implied Claim:}
Skagen Painter Peder Severin Krøyer favored naturalism along with Theodor Esbern Philipsen and the artist Ossian Elgström studied with in \textit{the early 1900s}.

\subsection{Claim Labeling}
The objective of this task is to identify the claims to be \textsc{Supported},\textsc{ Refuted}, or \textsc{NotEnoughInfo} given the supporting facts. 

\paragraph{Supported} 
You have strong reasons from the supporting documents, or based on your linguistic knowledge, to justify this claim is true. 

\paragraph{Refuted}
Based on the supporting documents, it's impossible for this claim to be true. You can find information contradicts the supporting documents in \textsc{ Refuted} claims. 

\paragraph{NotEnoughInfo}
Any claim that doesn't fall into one of the two categories above should be labeled as \textsc{NotEnoughInfo}. This usually suggests you need ADDITIONAL information to validate whether the claim is TRUE or FALSE after reviewing the paragraphs. Whenever you are not sure whether a claim is Refuted or \textsc{NotEnoughInfo}, ask yourself "Is it possible for this claim to be true based on the information from paragraphs?" If yes, select \textsc{NotEnoughInfo}.

\paragraph{External Knowledge.}
The concept of external knowledge is ambiguous and hard to define precisely, and the failure to address this issue could confuse workers regarding what information they are allowed to use when making their judgments.
To address this, we distinguish linguistic knowledge and commonsense from external, encyclopedia knowledge, as additional information that they are allowed to use in the task. 

Linguistic knowledge can be defined as vocabulary and syntax of an English speaker. It is invariant to most of the English speakers and can play a crucial role in this task. For example, given the supporting facts \textit{Messi is the captain of the Argentina national team.}, the claim was generated by substituting \textit{captain} to \textit{leader}. From our linguistic knowledge, \textit{captain} and \textit{leader} are synonyms, hence the mutated claim conveys the same idea as the provided supporting facts, and therefore should be annotated as \textsc{Supported}.   
On the other hand, if \textit{captain} is replaced by \textit{goalkeeper}, an English speaker can easily tell they are words of different meanings. 
Hence, additional information such as Messi's position should be provided in order to justify this claim. 
This type of information is beyond the supporting facts and should be considered as external information, and therefore the mutated claim should be annotated as \textsc{NotEnoughInfo}.
In addition to linguistic knowledge, commonsense should also be taken into account. 
Few examples of commonsense would be: a person can only have one birth place, a person cannot perform actions after their death, etc. Hence, claims which are found to not respect commonsense are labeled as \textsc{Refuted}.

\begin{table*}[t]

\begin{tabular}{p{7.5cm}p{7.5cm}}
\toprule
Paragraph 1:\textbf{  Northwestern University} & Paragraph 2:\textbf{ Middlebury College}\\
\hline
\textcolor{blue}{ Northwestern University (NU) is a private research university based in Evanston, Illinois, with other campuses located in Chicago and Doha, Qatar, and academic programs and facilities in Washington, D.C., and San Francisco, California.}

&
\textcolor{blue}{Middlebury College is a private liberal arts college located in Middlebury, Vermont, United States.} The college was founded in 1800 by Congregationalists making it the first operating college or university in Vermont...
\\

\hline
Paragraph 3: \textbf{Eddie George} & Paragraph 4:\textbf{ Hidden Ivies}\\

\hline
...\textcolor{blue}{Post-football, George earned an MBA from Northwestern University's Kellogg School of Management.} In 2016, he appeared on Broadway in the play ``Chicago" as the hustling lawyer Billy Flynn....
& 
\textcolor{blue}{Hidden Ivies: Thirty Colleges of Excellence is a college educational guide published in 2000.} It concerns college admissions in the United States....
\textcolor{blue}{In the introduction, the authors further explain their aim by referring specifically to ``the group historically known as the `Little Ivies' (including Amherst, Bowdoin, Middlebury, Swarthmore, Wesleyan, and Williams)" which the authors say ... }
\\
\hline
\end{tabular}
\begin{tabular}{p{15cm}}
\textit{\textbf{Claim:} The `Little Ivies', mentioned in the book Hidden Ivies, are Amherst, Bowdoin, Swarthmore, Wesleyan, Williams and one other. That other ``Little Ivy" and the institution where Eddie  George earned an MBA from, are both private schools in Pennsylvania.}\\
\end{tabular}

\begin{tabular}{p{7.5cm}p{7.5cm}}
\hline
\toprule

Paragraph 1:\textbf{  Flashbacks of a Fool} & Paragraph 2:\textbf{ Emilia Fox}\\

\hline
... \textcolor{blue}{The film was directed by Baillie Walsh, and stars Daniel Craig, Harry Eden, Claire Forlani, Felicity Jones, Emilia Fox, Eve, Jodhi May, Helen McCrory and Miriam Karlin.}
& 
... She also appeared as Morgause in the BBC’s ``Merlin" beginning in the programme's second series.\textcolor{blue}{She was educated at Bryanston School in Blandford, Dorset.}
\\
\hline
\end{tabular}

\begin{tabular}{p{15cm}}
\textit{\textbf{Claim:} 
Emilia Fox was a cast member of Flashbacks of a Fool was educated at Blandford Forum in Blandford, Dorset.}\\

\bottomrule
\end{tabular}

\caption{Two examples showing ambiguity between Refuted and NotEnoughInfo labels. In the first example, we need external geographical knowledge about Vermont, Illinois and Pennsylvania to refute the claim. In the second example, the claim cannot be directly refuted as Emilia Fox could have also been educated at Bryanston school and Blandford Forum.}
\label{table:amb}
\end{table*}

\subsubsection*{Instructions}

\begin{itemize}
    \item Review the claim. Then review the supporting documents, especially the highlighted sentences.
    \item Extract information from the supporting documents, to justify the given claim is \textsc{Supported} or \textsc{ Refuted}. If you are not certain and need additional information, please select \textsc{NotEnoughInfo}.
    \item Avoid using any external information that is not part of the supporting documents.
    \item If information from the claim and supporting documents is exclusive and is impossible to be both true, the claim should be labeled as \textsc{ Refuted}.
    \item If information from the claim and supporting documents is nonexclusive and it's possible that both can be true, the claim should be labeled as \textsc{ NotEnoughInfo}.

\end{itemize}

\paragraph{Examples of labeled claims}
Refer \tabref{tbl:task1ex-canada} for original claims, claim mutations and labels.

\paragraph{Refuted vs NotEnoughInfo.}
Refer \tabref{table:amb} for ambiguous examples.

\begin{table*}[!tbp]
\centering
\begin{tabular}{p{2cm}p{13.5cm}}

\hline
\textbf{Title} & \textbf{Wikipedia Article}\\
\hline

\textbf{Shanghai Noon
}  &

1. Shanghai Noon is a 2000 American-Hong Kong martial arts western comedy film starring Jackie Chan, Owen Wilson and Lucy Liu.\\
&
2. The first in the ``Shanghai (film series)".\\
&
\textcolor{blue}{3. The film, marking the directorial debut of Tom Dey, was written by Alfred Gough and Miles Mill}

\\
\hline 
\textbf{Tom Dey}

& 

\textcolor{blue}{1. Thomas Ridgeway ``Tom" Dey (born April 14, 1965) is an American film director, screenwriter, and producer.}\\
&
\textcolor{blue}{2. His credits include ``Shanghai Noon", ``Showtime", ``Failure to Launch", and ``Marmaduke".}
\\  
\hline 
\textbf{Roger Yuan}  & 
1. Roger Winston Yuan (born January 25, 1961) is an American Actor, martial arts fight trainer, action coordinator who trained many actors and actresses in many Hollywood films.\\
&
\textcolor{blue}{2. As an actor himself, he also appeared in ``Shanghai Noon" (2000) opposite Jackie Chan, ``Bulletproof Monk" (2003) alongside Chow Yun-fat, the technician in ``Batman Begins" (2005), and as the Severine's bodyguard in ``Skyfall" (2012).}\\
&
3: He is a well-recognized choreographer in Hollywood.
\\
\hline 
\textbf{Once Upon a Time in } &

\textcolor{blue}{1. Once Upon a Time in Vietnam (Vietnamese: Lua Phat ) is a 2013 Vietnamese action fantasy film directed by and starring Dustin Nguyen along with Roger Yuan.}\\\textbf{Vietnam} &
2. It was released on August 22, 2013.\\
&
3. This is the first Vietnamese action fantasy film.\\
\hline &
\textbf{2 hop Original Claim and  Claim Mutations}\\
\hline 
\textbf{Original} & Shanghai Noon was the directorial debut of an American film director whose other credits include Showtime, Failure to Launch, and Marmaduke. \textit{\textcolor{blue}{Supported}}\\  
\textbf{Entity Substitution } 
& Shanghai Noon was the directorial debut of a \textcolor{red}{Danish} film director whose other credits include Showtime, Failure to Launch, and Marmaduke. \textit{\textcolor{blue}{Not Supported}}\\
\hline &
\textbf{3 hop Original Claim and  Claim Mutations}\\
\hline 
\textbf{Original} & The film Roger Yuan appeared in was the directorial debut of an American film director. The director's other credits include Showtime, Failure to Launch, and Marmaduke. \textit{\textcolor{blue}{Supported}}\\ 
\textbf{More Specific} &
The film Roger Yuan \textcolor{red}{starred} in was the directorial debut of an American film director. The director's other credits include Showtime, Failure to Launch, and Marmaduke. \textit{\textcolor{blue}{Not Supported}}\\
\textbf{Entity Substitution} &

The film Roger Yuan appeared in was the directorial debut of an American film director. The director's other credits include Showtime, Failure to Launch, and \textcolor{red}{Steve Jaggi}. \textit{\textcolor{blue}{Not Supported}}\\
\hline &
\textbf{4 hop Original Claim and  Claim Mutations }\\
\hline 

\textbf{Original} & Roger Yuan starred in Once Upon a Time in Vietnam and another film that was the directorial debut of an American film director. The director's other credits include the Showtime, Failure to Launch, and Marmaduke. \textit{\textcolor{blue}{Supported}}\\ 
  
\textbf{Entity Substitution } & 
Roger Yuan starred in Once Upon a Time in Vietnam and another film that was the directorial debut of an American film director. The director's other credits include \textcolor{red}{the New York Times}, Failure to Launch, and Marmaduke. \textit{\textcolor{blue}{Not Supported}}

\end{tabular}
\caption{Original Claims, Mutated Claims with their supporting documents and labels.}
\label{tbl:task1ex-canada}
\end{table*}

\begin{figure*}[t]
    \centering
    \includegraphics[scale=0.42]{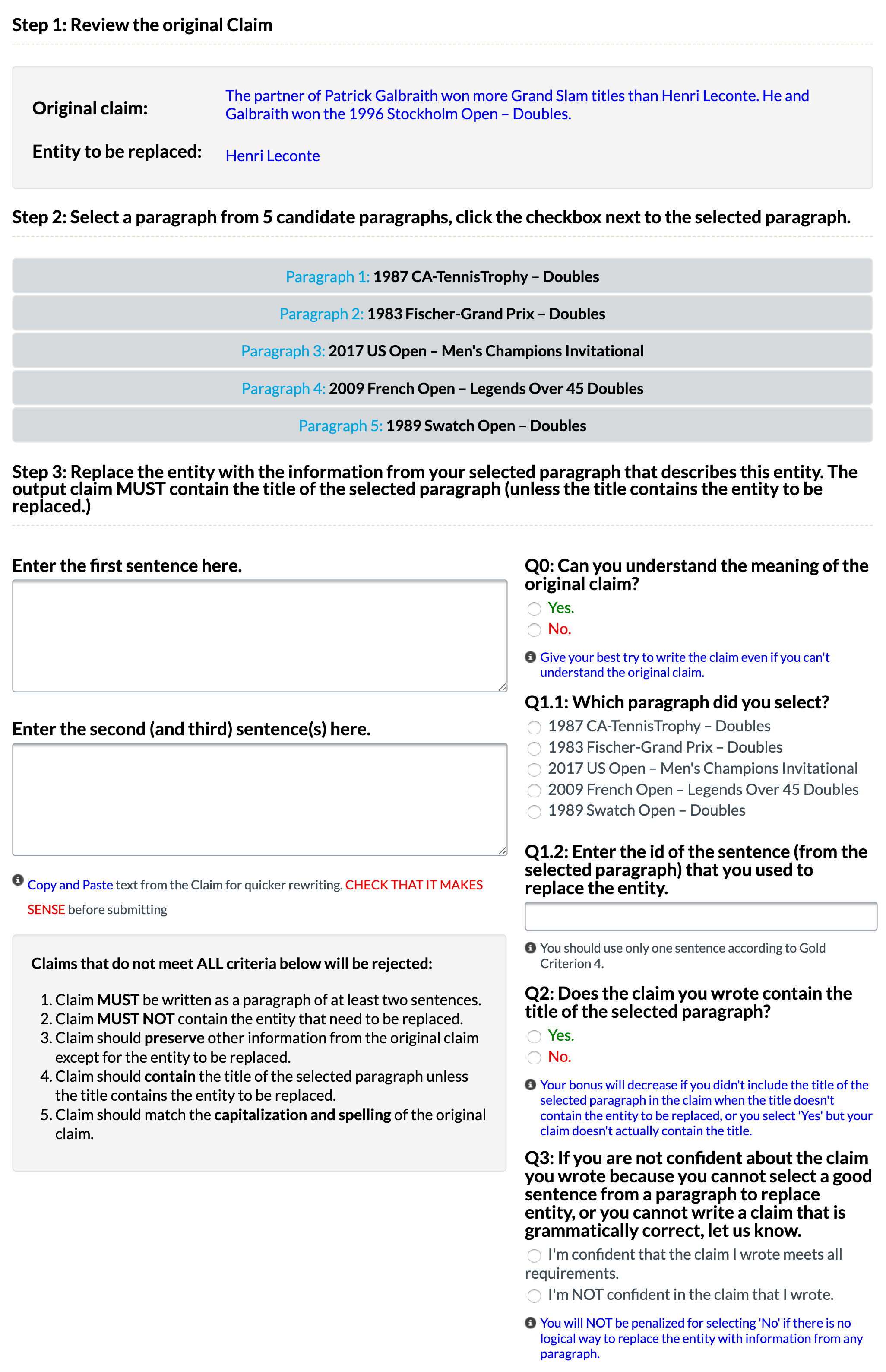}
    \caption{Screenshot of task to extend a 3-hop claim into a 4-hop claim.}
    \label{screenshot:hop_extension}
\end{figure*}

\begin{figure*}[t]  
    \centering
    \includegraphics[scale=0.45]{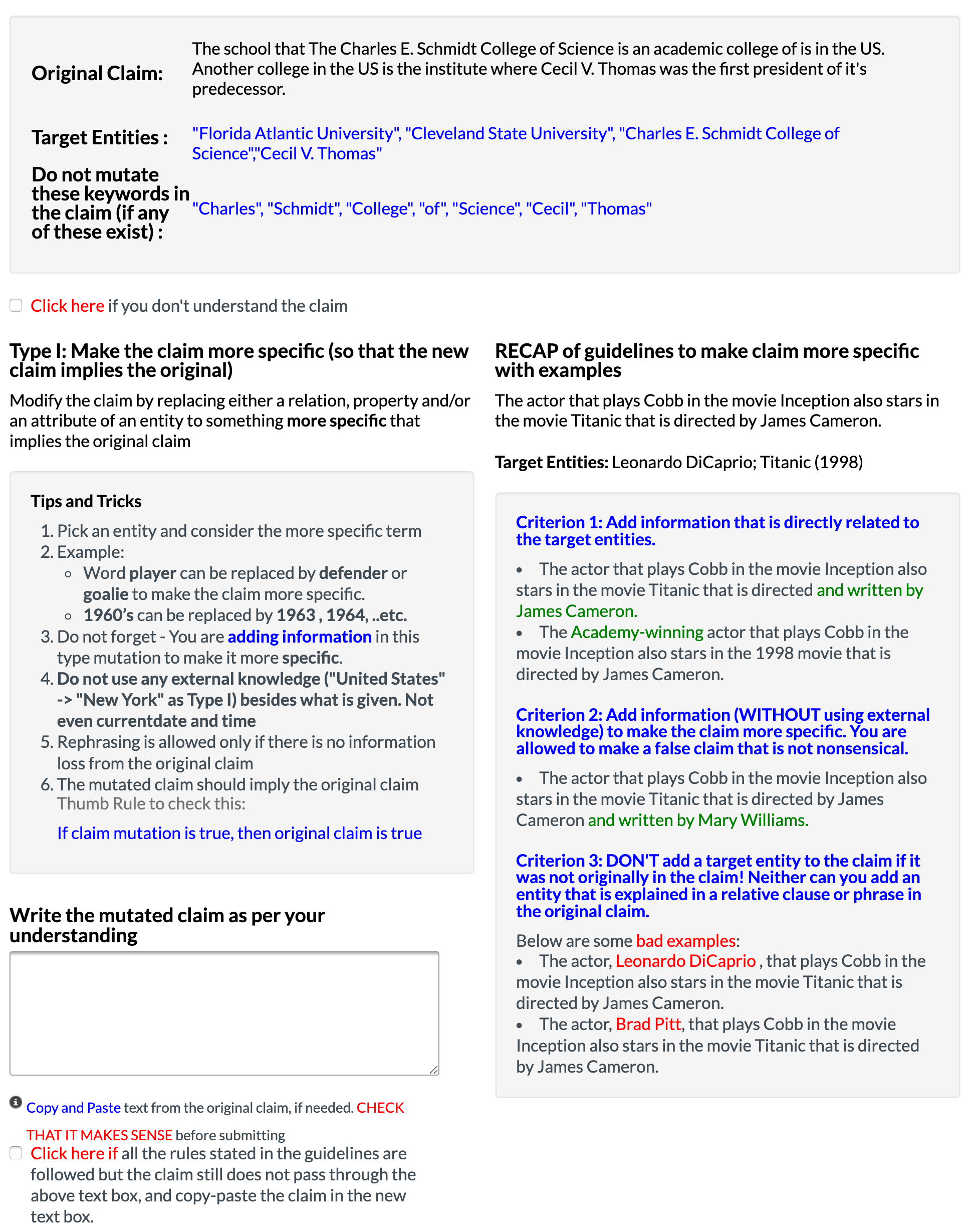}
    \caption{Screenshot of Creating More Specific Claims.}
    \label{screenshot:type5}
\end{figure*}

\begin{figure*}[t]
    \centering
    \includegraphics[scale=0.42]{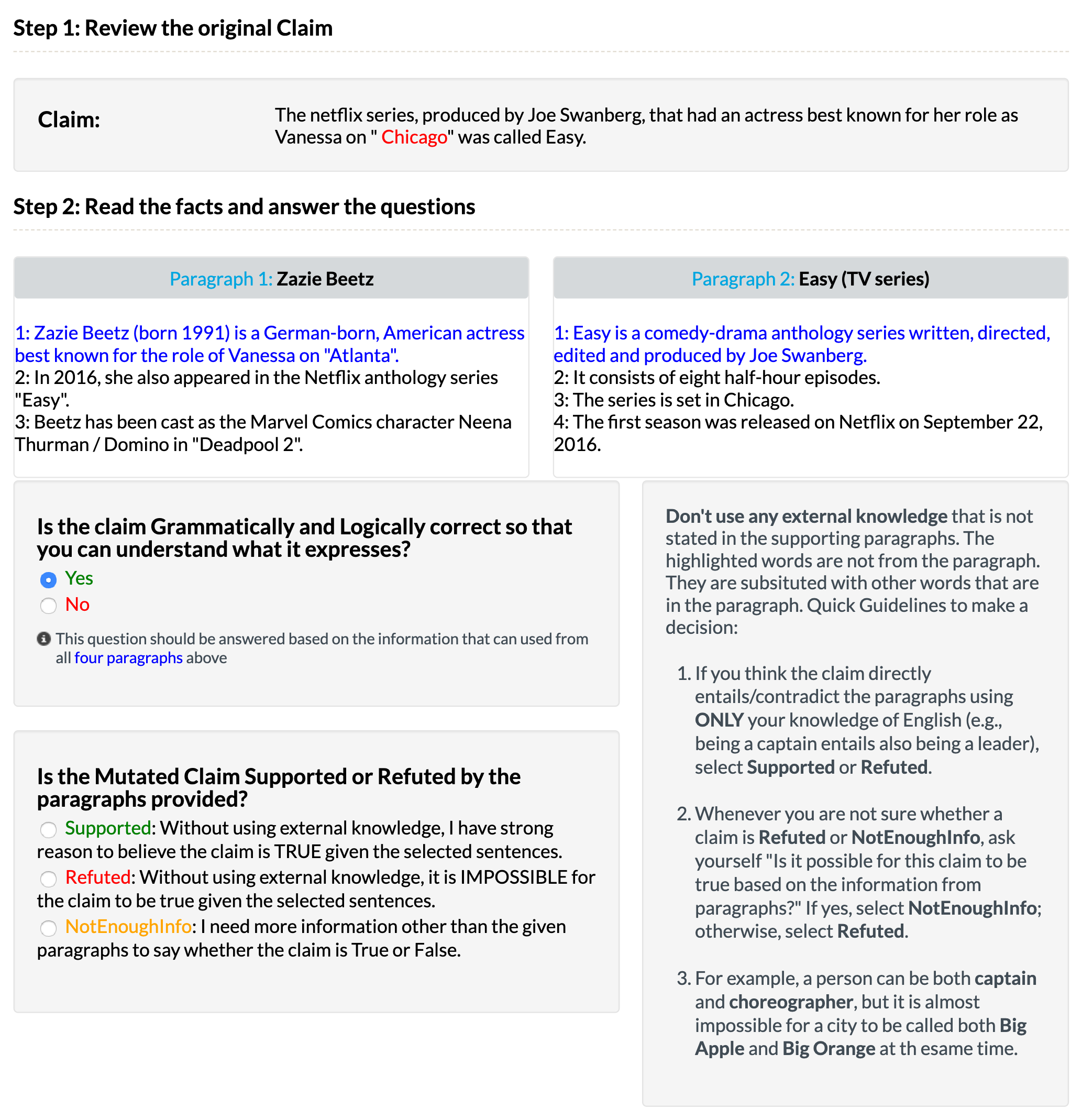}
    \caption{Screenshot of Labeling Task.}
    \label{screenshot:labeling}
\end{figure*}

\end{document}